# AECV-Bench: Benchmarking Multimodal Models on Architectural and Engineering Drawing Understanding

Authors: Aleksei Kondratenko, PhD§; Mussie Birhane§; Houssame E. Hsain‡; Guido Maciocci§

§ AECFoundry
‡ University of New South Wales, Kensington, NSW, Australia

https://github.com/AECFoundry/AECV-Bench

## Abstract

AEC drawings encode geometry and semantics through symbols, layout conventions, and dense annotation, yet it remains unclear whether modern multimodal and vision-language models can reliably interpret this graphical language. We present **AECV-Bench**, a benchmark for evaluating multimodal and vision-language models on realistic AEC artefacts via two complementary use cases: (i) object counting on 120 high-quality floor plans (doors, windows, bedrooms, toilets), and (ii) drawing-grounded document QA spanning 192 question–answer pairs that test text extraction (OCR), instance counting, spatial reasoning, and comparative reasoning over common drawing regions. Object-counting performance is reported using per-field exact-match accuracy and MAPE results, while document-QA performance is reported using overall accuracy and per-category breakdowns with an LLM-as-a-judge scoring pipeline and targeted human adjudication for edge cases. Evaluating a broad set of state-of-the-art models under a unified protocol, we observe a stable capability gradient; OCR and text-centric document QA are strongest (up to 0.95 accuracy), spatial reasoning is moderate, and symbol-centric drawing understanding – especially reliable counting of doors and windows – remains unsolved (often 0.40–0.55 accuracy) with substantial proportional errors. These results suggest that current systems function well as document assistants but lack robust drawing literacy, motivating domain-specific representations and tool-augmented, human-in-the-loop workflows for safe AEC automation.

## 1. Introduction

The Architectural Engineering and Construction (AEC) industry heavily relies on 2D drawings as the primary medium for conveying design intent, coordinating disciplines, and documenting as-built conditions. Architectural and engineering drawings, plans, sections, details, and specifications encode not only geometry, but also rich semantic information in the form of symbols, annotations, tags, and cross-references to specifications and standards. Reading, cross-checking, and updating this information consumes a large fraction of engineers' and architects' time, and errors in interpretation easily propagate directly into clashes, inaccurate estimates, rework, and safety risks.

Currently, multimodal large language models (MMLMs) and vision language models (VLMs) have made impressive progress on generic visual understanding (Yin et al., 2024; OpenAI, 2023; Gemini Team, 2023; Liu et al., 2023). Off-the-shelf APIs can now answer open-ended questions, read texts, extract information from images, scanned PDFs, and summarize complex documents. These advancements have sparked significant interest across the AEC industry; vendors and practitioners are actively experimenting with LLM-powered assistants that promise to extract information (Zheng & Fischer, 2023; Prieto et al., 2023), count doors and windows (Schönfelder et al., 2024; Li & Wang, 2024; Zhang & Zhou, 2023), check code compliance (Ying & Sacks, 2024; Chen et al., 2023), and automatically generate reports from design package (Chen & Fang, 2025; Liao & Zhou, 2024).

However, despite this enthusiasm, it remains unclear how well current models really understand AEC drawings. Most widely used benchmarks for multimodal and vision models focus heavily on perception-heavy tasks: vision question answering, captioning, OCR/document understanding, basic object/scene recognition, and relatively short-horizon reasoning grounded in single images or short clips (Fu et al., 2024; Mathew et al., 2021; Mathew et al., 2022; Van Landeghem et al., 2023; Masry et al., 2022). Even when diagrams or floor plans are included, they typically involve simplified synthetic layouts, a limited symbol vocabulary, or tasks that do not represent the day-to-day engineering work. In contrast, production drawings in real projects are messy; legacy CAD conventions, dense annotation, and discipline-specific symbols that require years of training to interpret correctly. This gap between marketing claims and practice motivates a more grounded question as to what extent do current multimodal and vision models actually understand real AEC drawings, as opposed to merely reading the text that happens to be printed on them?

To address this question, we introduce AECV-Bench, a new benchmark designed to evaluate "drawing intelligence" of multimodal and vision models on realistic AEC artefacts. The benchmark is built from a curated collection of high-quality floor plans and related documents sampled from real-world projects. Each drawing is manually analyzed and annotated to support several complementary families of questions that mirror common tasks in AEC workflows, including instance counting, text extraction (OCR), spatial reasoning, and comparative reasoning. We evaluated a diverse set of state-of-the-art models spanning proprietary frontier systems (e.g., Gemini, GPT, Claude, Grok) and strong open or smaller models (e.g., Qwen, GLM, Mistral, Cohere). All models receive the same prompts and scoring protocol, and answers are evaluated using a combination of automated matching and human-in-the-loop adjudication for ambiguous cases.

This paper makes the following contributions:

1. **Benchmark design for AEC drawing intelligence.** We introduce AECV-Bench, a multi-task evaluation suite for multimodal and vision models on real AEC drawings and related documents, covering object counting, OCR-based QA, spatial reasoning, and document QA.

2. **Curated dataset and annotation pipeline.** We describe a data collection and labeling process for high-quality floor plans and associated artefacts, including object-level counts, manually authored question–answer pairs, and a human-in-the-loop scoring protocol tailored to noisy and ambiguous real-world drawings.

3. **Comprehensive empirical evaluation.** We systematically evaluate a broad set of state-of-the-art models under a unified experimental setup, quantify strengths and weaknesses across task families, and highlight a clear performance gradient across OCR, spatial reasoning, and symbol-centric drawing understanding.

4. **Discussion of implications for AEC, model providers, and research.** Based on the observed results, we analyze what is currently safe and valuable to automate in AEC workflows, identify limitations of generic multimodal pretraining for technical drawings, and outline research directions in representation, architecture, and human – AI collaboration.

By grounding evaluation in the messy reality of AEC drawings, AECV-Bench aims to serve both as a practical decision-making tool for industry and as a challenging testbed for the broader multimodal research community.

## 2. Related Work

The rapid advancement of multimodal artificial intelligence has created unprecedented opportunities for automating document-intensive tasks across industries; yet significant gaps remain between demonstrated capabilities on standard benchmarks and the specialized requirements of technical domains such as Architecture, Engineering, and Construction (AEC).

**Multimodal Large Language Models and Vision-Language Models.** The evolution of vision-language models has accelerated dramatically since the introduction of CLIP (Radford et al., 2021), which demonstrated that contrastive pretraining on large-scale image-text pairs enables powerful zero-shot visual recognition. Subsequent architectures built upon this foundation: BLIP introduced bootstrapped caption generation for unified understanding and generation (Li et al., 2022), while BLIP-2 proposed the Q-Former architecture to efficiently bridge frozen image encoders with large language models (Li et al., 2023). The current generation of multimodal large language models (MMLMs), including GPT-4V (OpenAI, 2023), Gemini (Gemini Team, 2023), and open-weight alternatives such as LLaVA (Liu et al., 2023), extend these foundations through visual instruction tuning, enabling conversational interaction over images with emergent capabilities in reasoning, OCR, and document comprehension. A comprehensive survey by Yin et al. (2024) characterizes these systems across architecture, training paradigms, and emergent abilities, noting that while MMLMs excel at general visual question answering and natural image understanding, their performance degrades substantially on domain-specific technical content.

**Document Understanding and OCR Benchmarks.** Parallel progress in document understanding has yielded a suite of benchmarks that probe visual questions answering over textual artifacts. DocVQA (Mathew et al., 2021) established the paradigm of answering questions grounded in scanned documents, while TextVQA (Singh et al., 2019) extended this to scene text in natural images. Subsequent benchmarks have targeted increasingly complex document types: InfographicVQA (Mathew et al., 2022) requires joint reasoning over layout, graphics, and data visualizations; ChartQA (Masry et al., 2022) focuses on visual and logical reasoning about charts; and DUDE (Van Landeghem et al., 2023) provides multi-domain, multi-industry document understanding tasks revealing substantial gaps between model and human performance. Specialized architectures such as LayoutLM (Xu et al., 2020) incorporate explicit text-layout pretraining to improve document image understanding. However, these benchmarks predominantly feature natural documents—forms, receipts, reports, and infographics—rather than technical drawings with symbolic conventions, geometric relationships, and discipline-specific notation.

**Floor Plan Analysis and AEC Computer Vision.** Research on automated floor plan analysis has progressed from rule-based symbol recognition to deep learning approaches. The CubiCasa5K dataset (Kalervo et al., 2019) provided 5,000 annotated floor plans spanning over 80 categories, enabling multi-task learning for room segmentation and symbol detection. Pizarro et al. (2022, 2023) advanced automatic floor plan analysis through wall vectorization and panoptic symbol spotting, culminating in a large-scale multi-unit dataset. Object detection approaches have applied YOLO variants and custom architectures to identify doors, windows, and structural elements (Schönfelder et al., 2024; Li & Wang, 2024; Zhang & Zhou, 2023). A recent comprehensive survey by Khade et al. (2025) synthesizes two decades of floor plan analysis research, noting the transition from handcrafted features to convolutional and graph neural networks. Despite this progress, existing work focuses primarily on specialized detection and segmentation pipelines rather than evaluating general-purpose MMLMs on the full spectrum of drawing understanding tasks that practitioners face.

**AI and LLM Applications in the AEC Industry.** The construction industry has begun exploring large language models for information extraction and automation. Zheng and Fischer (2023) introduced BIM-GPT, a dynamic prompt-based framework for querying Building Information Models that achieved high accuracy on structured information retrieval. Researchers have investigated LLM-assisted code compliance checking (Ying & Sacks, 2024; Chen et al., 2023), finding that while models can transform regulatory text into computable rules, reliability remains insufficient for autonomous decisions. Recent reviews (Zhang & Zheng, 2025; Pan et al., 2024) catalog LLM applications across AEC project phases—design generation, document summarization, specification extraction—while cautioning that most deployments remain experimental. Critically, none of these efforts systematically benchmark general-purpose MMLMs on the reading and interpretation of production drawings, leaving practitioners without evidence-based guidance on what current systems can and cannot reliably accomplish on realistic AEC artifacts.

**Research Gap.** Taken together, the literature reveals a clear gap: while MMLMs demonstrate strong performance on natural images and standard document benchmarks, and while specialized computer vision pipelines address specific floor plan tasks, no comprehensive benchmark evaluates the full range of drawing intelligence, i.e. symbol interpretation, counting, spatial reasoning, and structured extraction, that would enable rigorous comparison of models on realistic AEC workflows. AECV-Bench addresses this gap by providing a multi-task evaluation suite, grounded in production-quality drawings and manually curated question-answer pairs that probe capabilities beyond text recognition.

## 3. Methodology

This section presents the complete methodology for AECV-Bench, covering data collection and annotation, benchmark design, execution pipelines, and evaluation procedures. AECV-Bench evaluates MMLMs and VLMs on two distinct use cases, each targeting different capabilities:

1. **Floor-plan object counting.** Given a floor plan image, the model counts doors, windows, bedrooms, and toilets. This task probes symbol recognition and instance enumeration capabilities (see Fig 1). The resulting benchmark dataset spans diverse drawing styles and annotation targets; representative examples (see Fig 2).

2. **Drawing-grounded document understanding (QA).** Given a drawing and a question, the model produces an answer spanning text extraction (OCR), spatial reasoning, instance counting, and comparative reasoning. This task evaluates broader document comprehension abilities. (see Fig 3).

Both pipelines share a unified API infrastructure for model invocation through OpenRouter API (supporting Claude, OpenAI GPT, Google Gemini, Grok, Mistral, Qwen, NVIDIA, Amazon Nova) and Cohere API. Standardized prompting and scoring protocols ensure fair comparison across all evaluated models.

We evaluate a range of models from two broad families. MMLMs (Multimodal Large Language Models) refer to general-purpose multimodal models designed for image and text reasoning via conversational prompting. VLMs (Vision-Language Models) refer to vision-first or vision-specialized models that support image-conditioned generation and/or visual question answering. The complete list of evaluated models is provided below, and we reference this list throughout the Results section.

- Gemini 3 Pro (Google)
- GPT-5.2 (OpenAI)
- Claude Opus 4.5 (Anthropic)
- Grok 4.1 Fast (xAI)
- Mistral Large 3 (Mistral AI)
- Qwen3 VL 8B Instruct (Alibaba / Qwen)[*]
- GLM-4.6V (Zhipu AI)
- NVIDIA Nemotron Nano 12B V2 VL (NVIDIA)
- Nova 2 Lite v1 (Amazon)
- Command A Vision (Cohere)

*(\*) Qwen3 is also available in substantially larger variants (e.g., Qwen3 235B A22B Instruct/Thinking). We report results for Qwen3-8B-Instruct because, in our evaluation setting, it offered the best speed–accuracy trade-off; larger variants introduced higher latency without commensurate gains for our tasks.*

### 3.1   Object Counting Pipeline

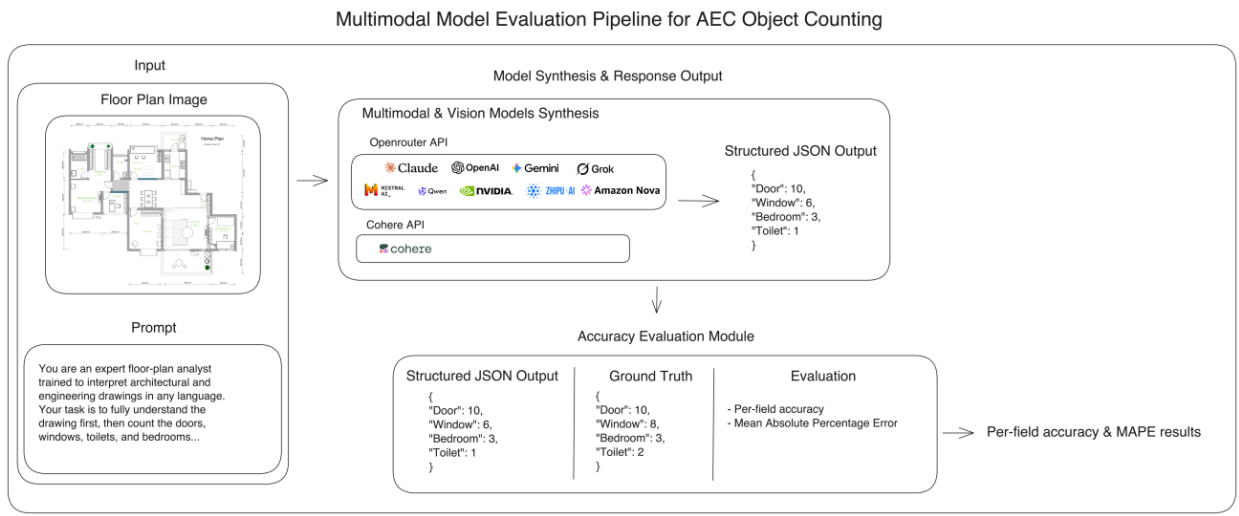

**Fig 1**: Floor-plan object counting pipeline: (1) Input floor plan image with structured counting prompt; (2) Model synthesis through OpenRouter API and Cohere API; (3) Structured JSON output with per-element counts; (4) Accuracy evaluation comparing predictions against ground truth; (5) Per-field accuracy and MAPE results across models.

#### 3.1.1   Data collection and annotation

For the object counting benchmark, we curated a dataset of 120 high-quality floor plans. These plans were assembled mainly from three sources. A subset was carefully selected from CubiCasa5K (Kalervo et al., 2019), and from the CVC-FP (de las Heras et al., 2015) datasets. The remaining plans were collected from publicly available drawings on the internet. No commercial collaboration was involved; all drawings were freely available at the time of collection.

Selection prioritized drawings that are representative of real AEC plan-reading conditions while remaining legible enough to support consistent annotation and evaluation. In particular, we favored floor plans with clear architectural conventions (such as walls, openings, and room delineations) and sufficient graphical fidelity to allow for reliable identification and counting of repeated instances.

Each floor plan was labelled with ground-truth counts for the following object classes:

- Doors
- Windows
- Bedrooms
- Toilets

Ground-truth labels were produced via manual review to ensure correctness under common plan ambiguities (e.g., doors drawn in different styles, windows represented by varying symbols, and bedrooms identified via room tags or typical layouts when tags are absent).

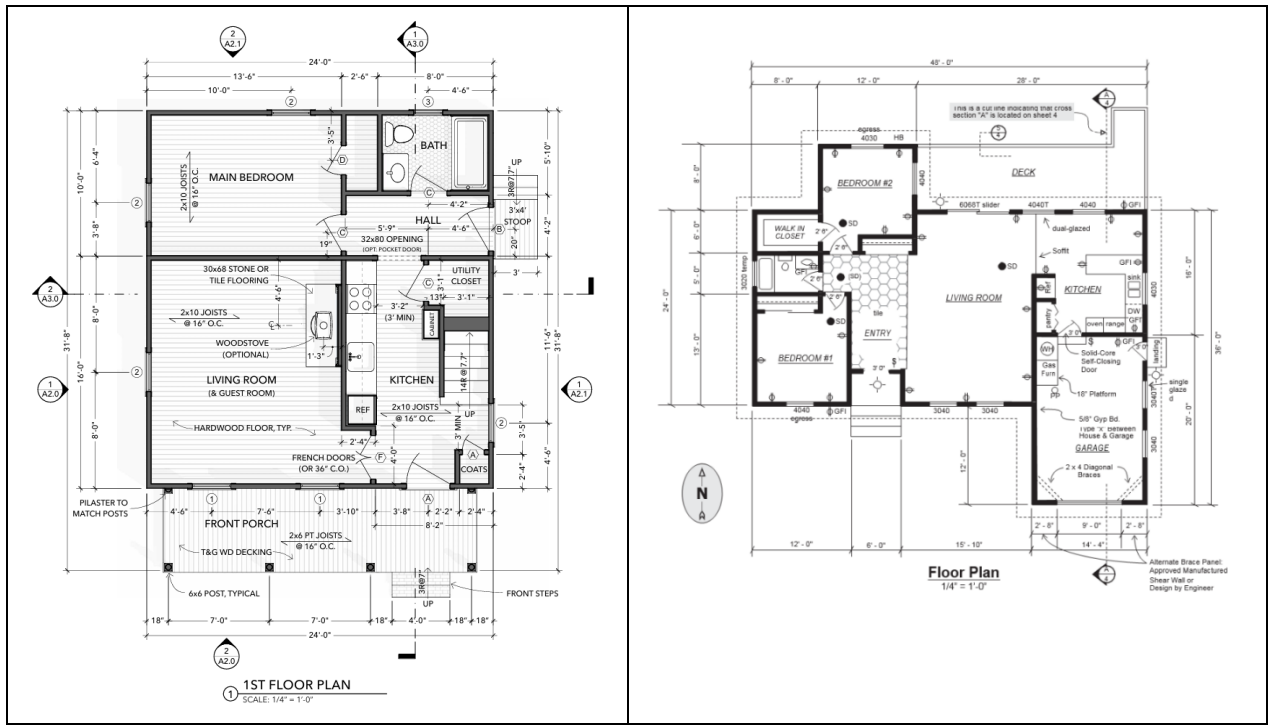

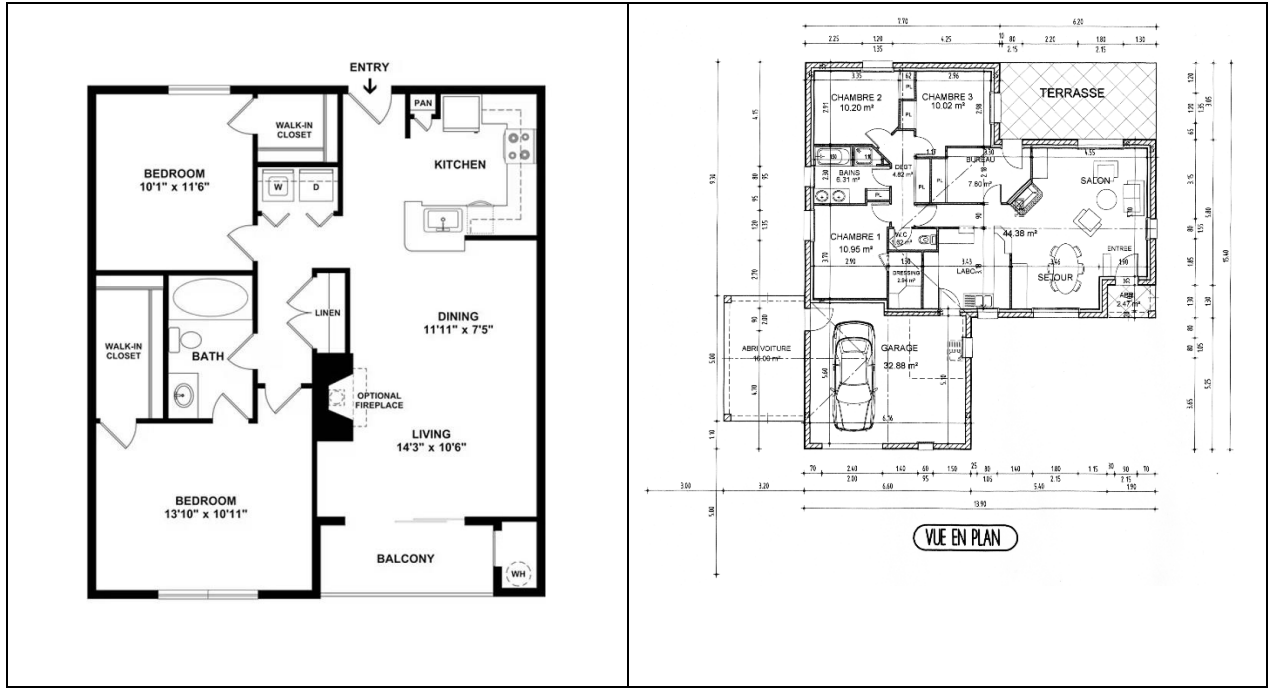

**Fig 2**: Dataset example

### 3.1.2 Task Formulation

*Task:* Given a floor plan image, the model is asked to output counts for doors, windows, bedrooms, and toilets.

*Prompting:* We used a single, strong, and consistent prompt for all models to minimize confounds due to prompt variation. The prompt instructs the model to examine the entire drawing and return the counts in a structured format. The prompt used is presented below:

```
You are an expert floor-plan analyst trained to interpret architectural and
engineering drawings in any language. Your task is to fully understand the
drawing first, then count the elements exactly according to the rules below and
then return ONLY the JSON strictly following the provided schema.

1. Counting Rules and Interpretation Guidelines
You must follow these rules precisely—these match the logic used in preparing
the ground-truth dataset. All counts MUST be based on the entire floor-plan
image.

1.1 General
- Always consider the entire drawing before counting anything.
- Understand the entire drawing as an architectural and/or engineering drawing
before counting anything.

Drawings may use different languages, symbols, and conventions; interpret them
consistently.

1.2 Doors
- Count only openable doors (swing doors drawn with an arc).
```

```
- Do NOT count sliding closet doors, pantry sliders, or furniture-like wooden
slides on internal walls.
- Each leaf counts separately, i.e., a double-leaf door = 2 doors and a triple-
leaf door = 3 doors, etc. and so on.
- Garage entrances are NOT doors.

1.3 Windows
- All sliding openings are counted as windows.
- French doors or large glazed openings are counted as windows, NOT doors.
- Multiple adjacent windows (or sliding units) without a gap between them count
as ONE single window group.
- Tiny toilet windows must be counted even if they are very small.
- Garage entrances are NOT windows.

1.4 Spaces / Rooms
- Count all enclosed spaces, even if unlabeled.
- Partnerships of adjacent but separate spaces (e.g., WC next to Bath) are
counted individually.

1.5 Bedrooms
A space labeled "Bedroom", "Chambre," "Zimmer," "Camera," etc., or equivalent
should be counted as a bedroom.

1.6 Toilets
- The following labels all count as toilet units, each counted individually:
WC, W.C., Bath, Bain, Douche, Shower, SDB, Bathroom, Toilet, etc.
- Lavatories are not counted as toilets.
Make sure your counts follow all the rules above.
```

*Structured outputs.* To ensure consistent evaluation across models, we required responses in a structured JSON format with fixed fields for each object class (doors, windows, bedrooms, toilets). This constraint reduces variation in free-form phrasing, simplifies automated parsing, and ensures that all models are evaluated on comparable outputs under the same schema.

*Metrics:* We evaluate object counting using two complementary views of performance. First, we report per-field exact-match accuracy (sometimes described as recall-style exactness), i.e., whether the predicted count matches the ground truth exactly for each class and aggregated across drawings. Second, we report Mean Absolute Percentage Error (MAPE) to quantify proportional counting error per class.

Accordingly, results are reported as per-field scores (Doors, Windows, Bedrooms, Toilets), aggregated "mean" scores across fields, and summarized using tables for interpretability.

### 3.1.3 Evaluation Metrics

We evaluate object counting using two complementary views of performance:

1. **Per-field exact-match accuracy** (sometimes described as recall-style exactness), i.e., whether the predicted count matches the ground truth exactly for each class and aggregated across drawings.

2. **Mean Absolute Percentage Error (MAPE)** to quantify proportional counting error per class.

Accordingly, results are reported as per-field scores (Doors, Windows, Bedrooms, Toilets), aggregated "mean" scores across fields.

### 3.2 Document QA Pipeline

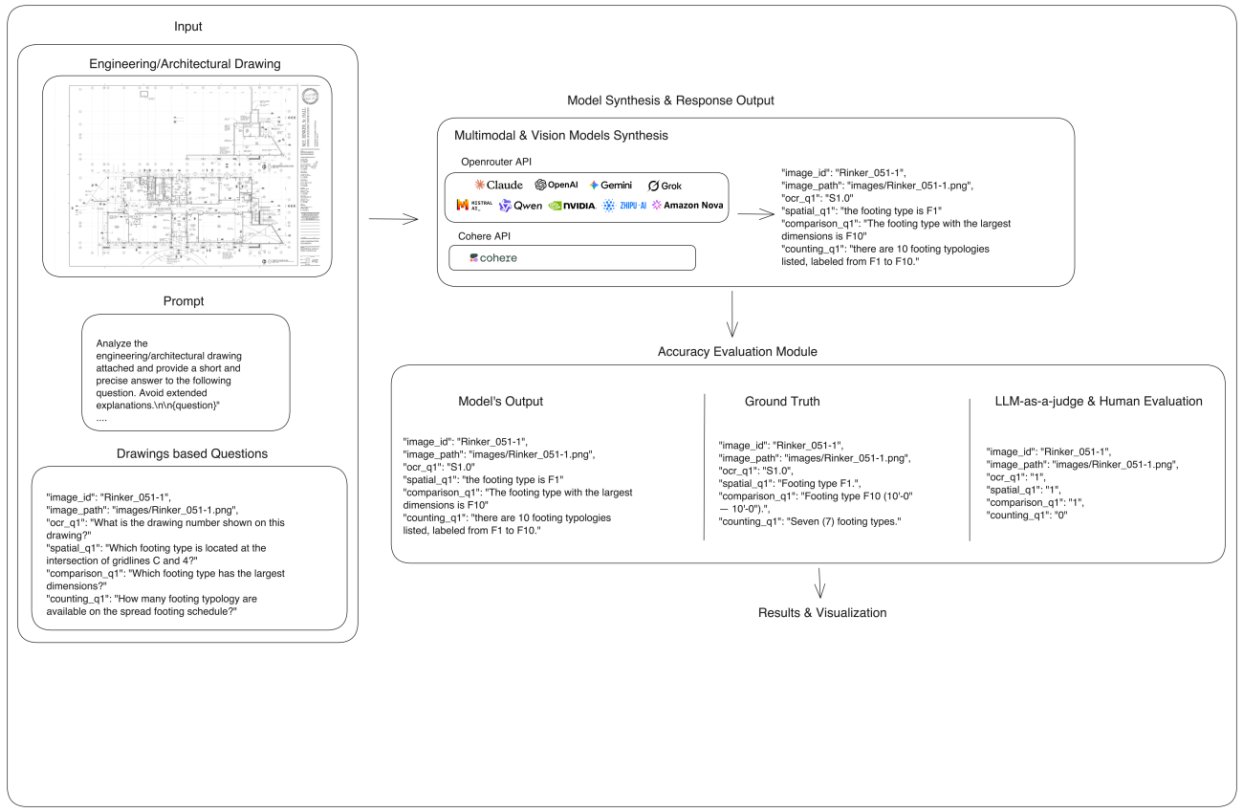

**Fig 3**: Document understanding QA pipeline: (1) Input engineering/architectural drawing with category-specific questions (OCR, spatial, counting, comparison); (2) Model synthesis through the same API infrastructure as object counting; (3) Model answers in structured format; (4) Accuracy evaluation via LLM-as-a-Judge with ground truth comparison; (5) Human adjudication for edge cases; (6) Visualization via overall and per-category bar charts.

#### 3.2.1 Data Collection and QA Authoring

For the QA benchmark, we manually crafted 192 question–answer pairs grounded in a collection of high-quality 21 drawings. QA pairs were written to probe multiple facets of drawing understanding beyond pure OCR, including spatial reasoning and symbolic interpretation, extending the scope of existing document QA benchmarks (Mathew et al., 2022; Masry et al., 2022) to technical drawing contexts.

The QA authoring and curation process were constructed through a combined model-assisted and human-curated workflow. We first used GPT-5 to propose candidate question–answer pairs for 21 drawings, targeting a broad range of drawing comprehension skills, including instance counting, spatial reasoning, comparative reasoning, text extraction (OCR), and symbol understanding, as well as information extraction from common drawing regions (e.g., title blocks, schedules, and tables). Human reviewers then audited these candidates, removed ambiguous or subjective items, verified that each question is answerable from the image alone, and selected a final set of 192 QA pairs. Reviewers also authored additional QA pairs directly from the drawings to better reflect representative AEC workflows and to ensure balanced coverage across task categories.

The QA set spans the following categories:

**Instance Counting Questions:** questions that test the ability to identify and enumerate architectural elements and non-textual graphics conventions, for example, counting the number of doors/windows in a floor plan, or identifying which rooms are served by a given stair. These tasks often require multiple steps: understanding what to count, locating the relevant view or region, recognizing instances of the element (including symbol interpretation), and reliably counting them.

**Text Extraction (OCR) Questions:** text-centric questions that test a model's ability to locate and accurately read labels, values, notes, dimensions, and metadata directly from drawings (e.g., room names, area values, sheet numbers, scales, or dates). These questions typically have a single definitive textual answer present somewhere in the drawing.

**Spatial Reasoning Questions:** questions that test spatial reasoning over the drawing, including adjacency, connectivity, relative position, containment, and access paths. Unlike OCR questions, the answer often cannot be directly read from a single text region and must be inferred from spatial relationships among elements.

**Comparative Reasoning Questions:** questions that test multi-element reasoning by requiring comparisons across several candidates (e.g., determining which room is largest, which space is closest to a reference, or which area has more openings). These tasks require identifying relevant elements, extracting comparable attributes, and selecting the element that satisfies a criterion.

Each QA pair is associated with a specific drawing image, and answers were written in a consistent form to support binary correctness evaluation (see Section 3.2.3).

```
{
    "image_id": "Apple-headquarter-plan-15-page-02",
    "image_path": "images/Apple-headquarter-plan-15-page-02.png",
    "ocr_qa": [
        {
            "id": "ocr1",
            "task": "scale_reading",
            "question": "What is the drawing scale indicated in the title block?",
            "answer": "1\" = 64'",
            "evidence": [
                "The 'Scale @ Arch E1' field in the title block reads '1\"=64''."
            ]
        }
    ],
    "spatial_qa": [
        {
            "id": "sp1",
            "task": "general_notes_location",
            "question": "Where is the General notes section located in the drawing?",
            "answer": "The top right corner of the drawing.",
```

```
            "evidence": [
                "The General notes section is located in the top right corner of the drawing."
            ]
        }
    ],
    "counting_qa": [
        {
            "id": "cnt1",
            "task": "count_section_callouts",
            "question": "How many section view callouts are present in the drawing?",
            "answer": "Three (3)",
            "evidence": [
                "Three section view callouts are present in the drawing. These are labeled as 'P-10.00', 'P-10.01/10.02', and 'P-10.03'"
            ]
        }
    ],
    "comparison_qa": [
        {
            "id": "cmp1",
            "task": "more_core_access_quadrant",
            "question": "Which quadrant of the circular building has more core access?",
            "answer": "Quadrant 1 or the top right corner.",
            "evidence": [
                "Quadrant 1 has more core access than the others."
            ]
        }
    ]
}
```

### 3.2.2  Task Formulation

*Task:* Given a drawing image and a question, the model produces an answer.

*QA categories:* Each QA pair belongs to one of the benchmark capability categories: OCR, instance counting, spatial reasoning and comparative reasoning. This enables both overall scoring and per-category breakdown analysis.

*Execution:* We run each model across the full QA set of 192 question–answer pairs using standardized prompting that provides the question, the associated drawing image, and answer-format constraints where applicable.

### 3.2.3  LLM-as-a-Judge Evaluation

Because drawing QA can include minor phrasing differences (e.g., synonyms, formatting, units) and because some drawings contain inherent ambiguities, we adopt a human-in-the-loop scoring pipeline that combines automated evaluation with targeted adjudication.

**Binary correctness scoring (0/1):** each model response is scored as **1** if the answer is correct (i.e., semantically matches the ground truth under defined equivalence rules), and **0** otherwise. This yields per-question correctness, which we aggregate into overall accuracy and per-category accuracy.

**LLM-as-a-judge evaluation:** To scale evaluation reliably, we use an LLM-as-a-judge procedure, following emerging practices in multimodal evaluation (Fu et al., 2024). Inputs to the judge include the question, the ground-truth answer, the model's predicted answer. The judge produces a binary label (0 or 1), yielding a consistent scoring signal across the full QA set and all evaluated models.

### 3.2.4 Human adjudication

We incorporate human review for edge cases, including ambiguous drawings (e.g., unclear symbols or occluded regions), borderline equivalences (e.g., multiple valid phrasings), and potential judge/model inconsistencies (see Fig 4). Human adjudication ensures the final evaluation reflects drawing-grounded correctness rather than judge idiosyncrasies and serves as a quality-control layer for this domain-specific benchmark.

Human adjudication is a key component of AECV-Bench for scaling the benchmark and tailoring evaluation to specific use cases. Automated matching alone can mis-score valid outputs due to acceptable variation (e.g., wording, formatting, units) and genuine ambiguity in technical artefacts. Reviewers therefore adjudicate borderline cases and calibrate task-specific acceptance criteria, enabling consistent scoring while adapting the pipeline to downstream workflows.

**Fig 4**: Web based QA evaluation pipeline schematic

## 4. Result and discussion

### 4.1 Result aggregation

Here the results of each use case are aggregated separately for floor-plan object counting and drawing-grounded document understanding (QA).

#### 4.1.1 Object counting

For each floor plan, we compare the model-predicted counts against ground truth for doors, windows, bedrooms, and toilets. We then aggregate performance across the full set of drawings into: (i) per-class exact-match accuracy (i.e., the fraction of plans for which the predicted count exactly matches the ground truth for a given class), and (ii) per-class MAPE (%) to capture proportional counting error.

In addition to per-class metrics, we report mean scores aggregated across the four classes. To support interpretability and model comparison, we visualize these metrics as presented in Table 1 and Table 2.

**Table 1**: Object counting per-field accuracy/recall for MMLMs & VLMs

| Model | Mean Accuracy | Door | Window | Bedroom | Toilet |
|---|---|---|---|---|---|
| *Proprietary Models* | | | | | |
|  Google Gemini 3 Pro | **0.51** | **0.39** | **0.34** | 0.89 | **0.82** |
|  OpenAI GPT-5.2 | 0.49 | 0.28 | 0.27 | **0.91** | 0.76 |
|  Claude Opus 4.5 | 0.42 | 0.16 | 0.16 | **0.91** | 0.76 |
|  Grok 4.1 Fast | 0.37 | 0.09 | 0.17 | 0.73 | 0.67 |
|  Amazon Nova 2 Lite v1 | 0.32 | 0.09 | 0.06 | 0.65 | 0.70 |
|  Cohere Command A Vision | 0.29 | 0.10 | 0.08 | 0.60 | 0.55 |
| *Open-source Models* | | | | | |
|  GLM-4.6V | 0.39 | 0.09 | 0.03 | 0.79 | **0.82** |
|  Qwen3-VL-8B Instruct | 0.39 | 0.09 | 0.10 | 0.76 | 0.81 |
|  NVIDIA Nemotron Nano 12B v2 VL | 0.38 | 0.12 | 0.18 | 0.70 | 0.75 |
|  Mistral Large 3 | 0.32 | 0.10 | 0.09 | 0.74 | 0.46 |

Table 2: Object counting per-field MAPE for MMLMs & VLMs

| Model | Mean MAPE (%) | Door (%) | Window (%) | Bedroom (%) | Toilet (%) |
|---|---|---|---|---|---|
| *Proprietary Models* | | | | | |
|  Google Gemini 3 Pro | **16.0** | **15.0** | **20.4** | 4.8 | **11.9** |
|  OpenAI GPT-5.2 | 19.4 | 20.0 | 24.7 | 4.6 | 18.1 |
|  Claude Opus 4.5 | 24.9 | 30.7 | 37.3 | **2.8** | 21.1 |
|  Grok 4.1 Fast | 28.4 | 35.1 | 42.8 | 14.6 | 23.6 |
|  Amazon Nova 2 Lite v1 | 38.4 | 39.9 | 51.7 | 14.3 | 57.3 |
|  Cohere Command A Vision | 50.8 | 69.0 | 61.4 | 33.4 | 42.1 |
| *Open-source Models* | | | | | |
|  GLM-4.6V | 29.5 | 41.1 | 40.8 | 13.4 | 18.5 |
|  Qwen3-VL-8B Instruct | 30.5 | 46.2 | 53.5 | 11.3 | 13.6 |
|  NVIDIA Nemotron Nano 12B v2 VL | 35.4 | 50.7 | 69.3 | 15.0 | 13.2 |
|  Mistral Large 3 | 39.2 | 51.8 | 62.4 | 26.6 | 22.1 |

Even the strongest models plateau around 0.4–0.55 mean exact-match accuracy, with large and systematic errors for symbol-heavy classes such as doors and windows. This is evident in Table 1 and Table 2, where door/window exact-match accuracy is consistently low and MAPE remains high across models. Across the tested MMLMs, Gemini 3 Pro is the strongest overall performer (0.51 mean accuracy; 16.0% mean MAPE), followed closely by GPT-5.2 (0.49 mean accuracy; 19.4% mean MAPE). Among VLMs, GLM-4.6v leads (0.39 mean accuracy; 29.5% mean MAPE). However, even these top models achieve only modest door/window exact-match rates (e.g., Gemini: 0.39 door, 0.34 window), confirming that the main bottleneck is symbol interpretation and instance enumeration rather than general image–text reasoning.

In terms of open-source vs. commercial performance, proprietary frontier models remain ahead on average, particularly on the most symbol-dependent classes (doors/windows). The strongest open-source model in our object-counting set, GLM-4.6v, reaches 0.39 mean accuracy (with 29.5% mean MAPE), placing it in a competitive mid-tier—often comparable to or stronger than several commercial "lite" vision systems—yet still materially behind the best proprietary models on the hardest counts.

Models frequently misinterpret door swings, confuse windows with openings or façade elements, hallucinate fixtures, or miss instances in dense regions—failure modes that echo challenges documented in specialized floor plan analysis research (Khade et al., 2025; Pizarro et al., 2022). Importantly, these errors persist even when the same models perform well on OCR for the very same drawings, highlighting that the bottleneck lies not in reading text but in understanding the graphical language of AEC plans.

### 4.1.2 Drawing-grounded document understanding (QA)

For the QA benchmark, each model response receives a binary correctness score (0/1) from the LLM-as-a-judge pipeline with human adjudication (see Section 3.2.4). We aggregate these binary outcomes into overall accuracy per model across all QA pairs (see Fig 5), as well as accuracy by QA category (see Fig 6). Results are summarized using bar charts to enable direct comparison across models both overall and per-category.

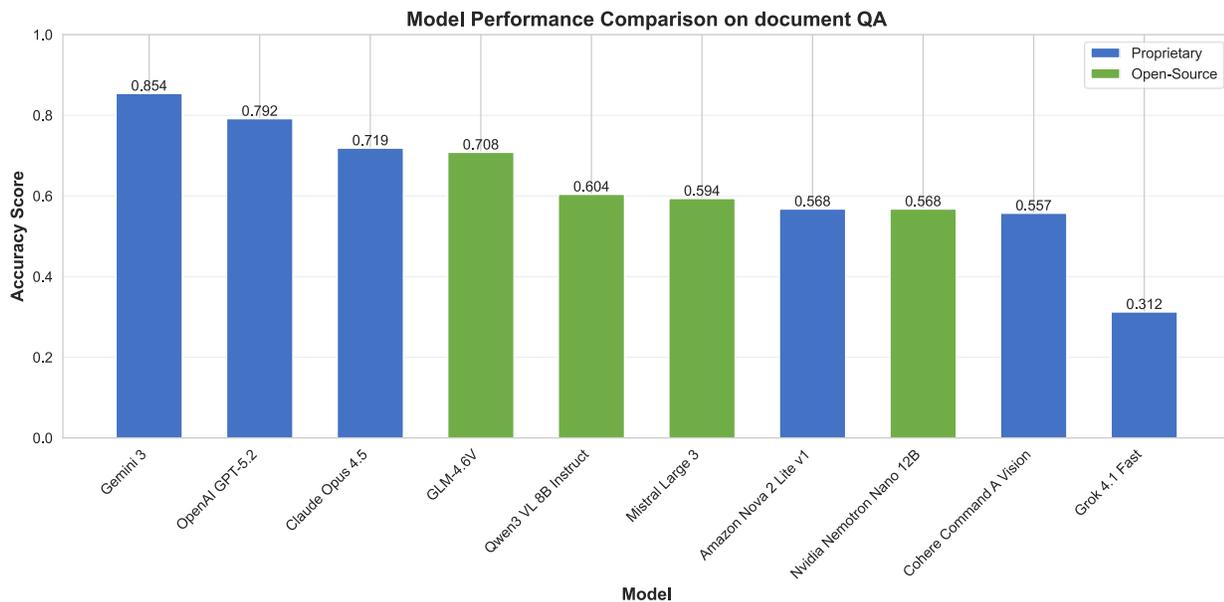

**Fig 5**: QA overall performance bar chart

Our experiments reveal a clear hierarchy of capabilities across the two AECV-Bench use cases: drawing-grounded document QA and floor-plan object counting. Several frontier models achieve strong overall accuracy, with the best-performing systems reaching the 0.70–0.95 range. As shown in Fig 5 and Fig 6, Gemini 3 leads the benchmark, followed by a cluster of large proprietary and strong open/vision models that achieve competitive performance. This aligns with the broader trajectory of multimodal research: when tasks primarily require locating, extracting, and reasoning over textual and lightly structured information in drawings, modern MMLMs and VLMs behave as capable document assistants, consistent with documented capabilities on standard document understanding benchmarks (Mathew et al., 2021; Van Landeghem et al., 2023).

A more granular breakdown by QA type reveals a consistent gradient across skills. Text extraction (OCR) tends to be the strongest category across top models (0.7-0.95), while spatial reasoning is reliably mid-range (0.6 to high-0.7), and instance counting remains the most error-prone component of the QA suite. Comparative reasoning typically lands between OCR and spatial reasoning, suggesting that models can compare extracted quantities when the underlying reads are reliable, but struggle when comparison requires robust spatial grounding or counting. This pattern is summarized in Fig 5 and Fig 6.

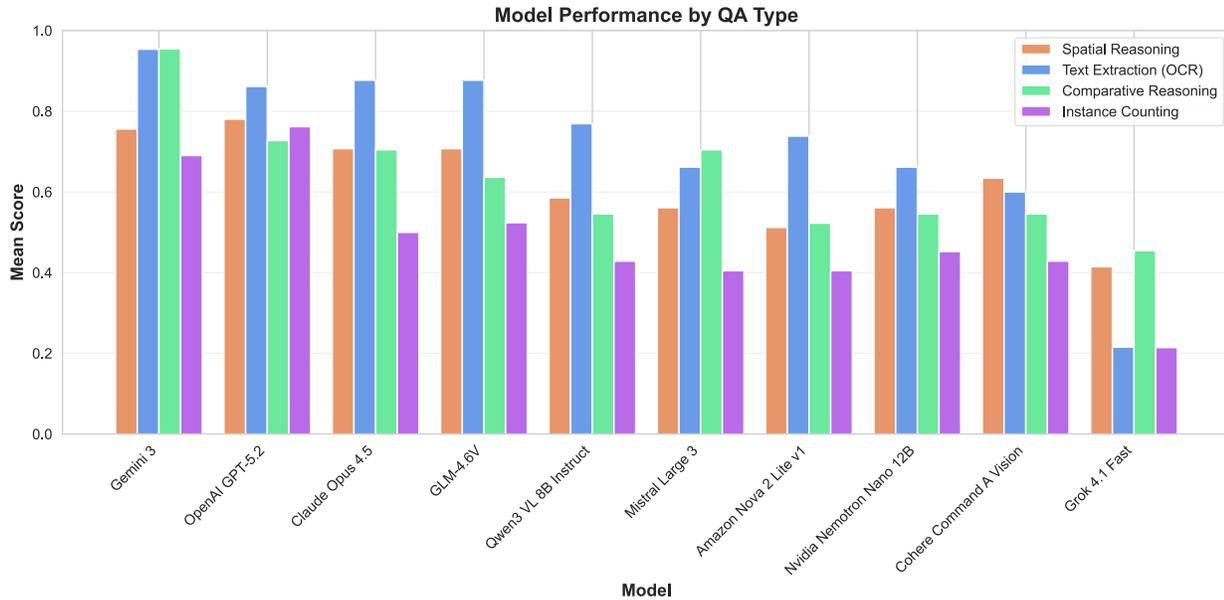

**Fig 6**: QA performance by QA type

### 4.2 Implications for the AEC Industry

From an industry perspective, our results argue for a selective and layered adoption of multimodal models rather than blanket automation of "drawing understanding".

First, the relatively strong performance on the document QA benchmark – particularly text extraction (OCR) and several comparative questions – supports the deployment of AI systems as productivity amplifiers around drawings and related documents.

Examples include:

- Faster Submittal Review by screening large drawing sets against specs to surface likely conflicts and omissions for reviewer sign-off;
- Quicker RFI triage and routing by summarizing drawing issues and directing them to the right discipline owner with relevant sheet context;
- Earlier contract risk visibility by checking scope obligations/exclusions against the included drawing set;
- Higher QA/QC throughput by flagging likely inconsistencies (e.g., missing tags, mismatched callouts) before formal review.

In these workflows, models handle most high-throughput reading and organization (around 90%), enabling human reviewers to spend their time on the 10% of edge cases and decisions that need expert judgment.

Second, the non-trivial performance on spatial reasoning suggests that models can already support assisted navigation and review of drawings. For example, they can help users quickly locate rooms between specified grids, highlight candidate spaces matching certain criteria, or provide approximate answers to "*where is…?*" queries. However, the current accuracy levels and fragility constrain these use cases to exploratory and supportive roles. Spatially-aware assistants should not be relied upon as the sole source of truth for layout-critical analyses such as fire-egress verification or travel-distance checks.

In contrast, the object-counting results on floor plans indicate that symbol-grounded counting (e.g., doors and windows) remains unreliable for automation without tight human supervision or custom trained models. This is especially important for workflows where counts flow directly into estimates, schedules, or compliance checks. In the near term, counting-oriented features should be deployed as assistive tools that surface candidates and uncertainty, rather than as autonomous extractors.

At the same time, not all counting targets are equally difficult; categories that are typically anchored by textual cues – notably bedrooms and toilets – tend to perform better than doors and windows. A plausible explanation is that many toilet and bedroom instances are room-based concepts that are frequently explicitly labeled (e.g., "*Toilet*," "*Bath*", "*W.C*", "*Bedroom*," room tags), allowing models to fall back on OCR and tag-based reasoning. By contrast, doors and windows are often conveyed primarily through graphical symbols and line-art conventions (swings, breaks in walls, façade notations) that vary across CAD standards and styles, making them harder to recognize and count consistently from rasterized drawings.

For AEC firms, the practical takeaway is that near-term value lies in automating text-centric and low-risk tasks, while treating any claims of full graphical understanding with caution. AI solutions should be designed as human-in-the-loop systems, with interfaces that surface uncertainties, allow efficient correction, and track model performance over time. Organizations that adopt such principled, benchmark-driven strategies are more likely to achieve real productivity gains without compromising safety or compliance.

## 4.3 Implications for Model Providers

For model providers, AECV-Bench exposes a significant domain gap between generic visual understanding and the requirements of technical drawings, consistent with survey findings that current MMLMs underperform on specialized technical content (Yin et al., 2024).

First, the strong performance on document QA co-existing with weak performance on floor-plan object counting suggests that current pretraining regimes heavily emphasize natural images and text, with relatively little exposure to vector-like artefacts such as CAD plans, engineering diagrams, or schematics. Closing this gap will likely require:

- Incorporating large-scale, diverse collections of technical drawings and CAD exports into pretraining corpora;
- Explicitly modelling line art, symbols, and geometric relationships, possibly through hybrid raster–vector representations;
- Designing architectures or training objectives that favors topological and symbolic consistency rather than purely local pattern recognition.

Notably, a number of AEC-focused vendors like Kreo (Kreo Software, 2026) and Togal (Togal AI, 2024) have introduced specialized drawing-understanding products that target tasks such as takeoff and element recognition. However, their reported capabilities are rarely benchmarked under transparent, standardized protocols, making it difficult to quantify performance, compare across systems, or understand failure modes. Benchmarks such as AECV-Bench help fill this gap by providing a reproducible evaluation setup for technical drawing intelligence.

Second, the results motivate the development of domain-aware interfaces and APIs. Many AEC workflows do not need open-ended text responses; they need structured outputs such as lists of rooms, door inventories, adjacency graphs, or grid mappings. Exposing such capabilities as first-class outputs—either through specialized "tools" within a general model or via domain-specific models, would make integration into production systems substantially easier and safer than relying on free-form answers.

Third, evaluation practices for multimodal models should evolve to reflect domain-specific requirements. Reporting a single "vision score" or performance on generic captioning datasets is insufficient for regulated or safety-critical domains. Benchmarks like AECV-Bench highlight the importance of:

- Decomposing evaluation into task-specific capabilities (text extraction (OCR), spatial reasoning, comparative reasoning, and instance counting);
- Publishing per-task and per-domain breakdowns alongside aggregate metrics;
- Clearly communicating failure modes and recommended use cases in documentation, particularly when marketing models to industries such as AEC.

Providers who invest in targeted training and transparent evaluation for technical domains stand to differentiate themselves meaningfully from generic offerings, especially as enterprises begin to demand evidence of capability before integrating AI into critical workflows.

**4.4   Implications for the Research Community**

From a research standpoint, AECV-Bench positions AEC drawings as a compelling testbed for multimodal reasoning that goes beyond natural images, addressing calls in recent reviews for rigorous evaluation of LLMs in AEC contexts (Zhang & Zheng, 2025; Pan et al., 2024).

First, the benchmark illustrates that technical drawings require a form of understanding that is inherently structured and symbolic. Rooms, doors, windows, and grids form graphs of entities and relations rather than collections of independent pixels. This naturally motivates exploration of:

- Neuro-symbolic architectures that parse drawings into explicit graphs or scene representations and perform reasoning over these structures;
- Vector-native and graph-based encoders that ingest CAD or IFC directly, complementing or replacing rasterization;
- Compositional training objectives that encourage models to maintain global consistency in counts, connectivity, and topology.

Second, the performance gradients across document QA categories (OCR, spatial reasoning, comparative reasoning, and instance counting) highlight the limitations of treating multimodal reasoning as a single monolithic capability. Future work can exploit this structure by:

- Designing task-specific adapters or modules—for example, coupling dedicated symbol detectors (Schönfelder et al., 2024; Li & Wang, 2024) with general MMLMs;
- Exploring tool-augmented agents that combine vision models with external algorithms, such as connected-component analysis, geometric solvers, or automated code-checking engines (Ying & Sacks, 2024; Chen et al., 2023);
- Leveraging human corrections in real workflows for active learning and continual improvement.

Finally, the safety-critical nature of many AEC decisions raises important questions about evaluation, uncertainty, and governance that extend beyond this domain.

Research is needed on:

- Quantitative methods for estimating confidence and calibrating outputs in multimodal settings;
- Principled ways to combine probabilistic model predictions with conservative rule-based systems;
- User interfaces and processes that enable engineers to remain "in the loop" without being overwhelmed by verification tasks.

## 5. Conclusion

Overall, our findings paint a nuanced picture of the current state of multimodal AI for AEC. Existing models are already powerful enough to deliver substantial value as document and information assistants around drawings, and their capabilities in OCR and simple layout reasoning are rapidly improving. At the same time, they fall short of genuine drawing literacy: understanding symbols, geometry, and code-relevant semantics in a manner robust enough for autonomous decision-making.

Across both pipelines, the key pattern is consistent: models are relatively strong at document-style understanding (especially OCR and related reasoning), but remain weak at symbol-grounded floor-plan counting, which is central to genuine drawing literacy.

Benchmarks such as AECV-Bench make these strengths and limitations visible. For industry, they help align expectations and guide investment toward realistic use cases. For model providers, they indicate where additional domain-specific innovation is required. For researchers, they open a rich space of problems at the intersection of vision, language, geometry, and formal reasoning.

## 6. Limitations

Several limitations of this study should be acknowledged when interpreting results and applying findings to practice.

**Dataset size and diversity.** The benchmark comprises 120 floor plans for object counting and 192 question-answer pairs for drawing QA. While sufficient for comparative evaluation, this scale limits the statistical power for fine-grained analyses and may not capture the full diversity of drawing styles, symbologies, and discipline-specific conventions encountered across the global AEC industry.

**Focus on 2D raster images.** All evaluations use rasterized PNG images of drawings rather than native CAD or BIM formats (DWG, RVT, IFC). This design choice reflects common real-world scenarios where practitioners work with scanned or exported drawings, but it precludes evaluation of models that could leverage vector geometry or semantic object hierarchies available in structured formats.

**Single-image evaluation.** Each benchmark item presents a single drawing image without cross-referencing to other sheets, details, or specifications. Real AEC workflows frequently require multi-page reasoning – for example, tracing a section callout to its corresponding detail sheet or reconciling room schedules across plan sets, which this benchmark does not assess.

**Annotator subjectivity.** Despite quality control measures, ground-truth counts and QA answers reflect the judgments of a limited annotator pool. Ambiguous cases, such as doors drawn in atypical styles or rooms identifiable only by context, introduce potential bias that may favor or penalize certain model behaviors.

**Limited object classes.** The counting task addresses only four object classes (doors, windows, bedrooms, toilets). Many other elements critical to AEC workflows, i.e. stairs, elevators, mechanical equipment, structural members, fire-rated assemblies, are not evaluated, limiting generalizability to broader drawing interpretation tasks.

## 7. Future work

Building on the findings and limitations of this study, we outline several directions for future development of AECV-Bench and related efforts.

**Industry partnerships for practice-grounded evaluation.** We aim to establish collaborations with AEC industry partners to access large-scale collections of production drawings paired with real practitioner queries and workflows. Such partnerships would enable evaluation scenarios grounded in actual professional needs rather than researcher-authored questions, yielding results with direct applicability to practice.

**Continuous leaderboard updates.** As new multimodal models are released by research labs and commercial providers, we plan to maintain and regularly update public leaderboard on our website (AECFoundry, 2021). This living benchmark will allow the community to track progress over time and provide model developers with timely feedback on AEC-specific performance.

**Accessible tools for AEC professionals.** Recognizing that current models already deliver value on OCR and document-centric tasks, we intend to develop simple, easy-to-use products that enable AEC professionals to leverage these capabilities without requiring technical expertise. By packaging proven strengths into intuitive interfaces, we hope to accelerate practical adoption while the research community continues to address harder challenges.

**Domain-specific model development.** Finally, we are exploring the development of specialized models targeting the symbol understanding gap identified in our results. By combining domain-specific pretraining on AEC drawings with architectural knowledge, we aim to close the performance gap between text-based and graphical reasoning tasks, ultimately enabling more robust and trustworthy AI assistance for drawing interpretation.

## 8. Data Availability Statement

All source code, accompanying files, and data generated and developed during this study are openly available in the following repository: https://github.com/AECFoundry/AECV-Bench